\documentclass[conference]{IEEEtran}
\IEEEoverridecommandlockouts

\usepackage{cite}
\usepackage{amsmath,amssymb,amsfonts}
\usepackage{algorithmic}
\usepackage{graphicx}
\usepackage{textcomp}
\usepackage{multirow}
\usepackage{booktabs}
\usepackage[table,xcdraw]{xcolor}
\usepackage[normalem]{ulem}
\useunder{\uline}{\ul}{}
\def\BibTeX{{\rm B\kern-.05em{\sc i\kern-.025em b}\kern-.08em
    T\kern-.1667em\lower.7ex\hbox{E}\kern-.125emX}}
\begin{document}

\title{MamMIL: Multiple Instance Learning for Whole Slide Images with State Space Models

\thanks{$^\ast$ Corresponding authors. This research received support from the National Natural Science Foundation of China (62031023 \& 62331011), and the Shenzhen Science and Technology Project (GXWD20220818170353009).}
}

\author{\IEEEauthorblockN{1\textsuperscript{st} Zijie Fang}
\IEEEauthorblockA{\textit{Tsinghua Shenzhen International Graduate School} \\
\textit{Tsinghua University}\\
Shenzhen, China \\
fzj22@mails.tsinghua.edu.cn}
\and
\IEEEauthorblockN{2\textsuperscript{nd} Yifeng Wang}
\IEEEauthorblockA{\textit{School of Science} \\
\textit{Harbin Institute of Technology (Shenzhen)}\\
Shenzhen, China \\
wangyifeng@stu.hit.edu.cn}
\and
\IEEEauthorblockN{3\textsuperscript{rd} Ye Zhang}
\IEEEauthorblockA{\textit{School of Computer Science and Technology} \\
\textit{Harbin Institute of Technology (Shenzhen)}\\
Shenzhen, China \\
zhangye94@stu.hit.edu.cn}
\and
\IEEEauthorblockN{4\textsuperscript{th} Zhi Wang$^\ast$}
\IEEEauthorblockA{\textit{Tsinghua Shenzhen International Graduate School} \\
\textit{Tsinghua University}\\
Shenzhen, China \\
wangzhi@sz.tsinghua.edu.cn}
\and
\IEEEauthorblockN{5\textsuperscript{th} Jian Zhang}
\IEEEauthorblockA{\textit{School of Electronic and Computer Engineering} \\
\textit{Shenzhen Graduate School, Peking University}\\
Shenzhen, China \\
zhangjian.sz@pku.edu.cn}
\and
\IEEEauthorblockN{6\textsuperscript{th} Xiangyang Ji}
\IEEEauthorblockA{\textit{Department of Automation} \\
\textit{Tsinghua University}\\
Beijing, China \\
xyji@tsinghua.edu.cn}
\and
\IEEEauthorblockN{7\textsuperscript{th} Yongbing Zhang$^\ast$}
\IEEEauthorblockA{\textit{School of Computer Science and Technology} \\
\textit{Harbin Institute of Technology (Shenzhen)}\\
Shenzhen, China \\
ybzhang08@hit.edu.cn}
}

\maketitle

\begin{abstract}
Recently, pathological diagnosis has achieved superior performance by combining deep learning models with the multiple instance learning (MIL) framework using whole slide images (WSIs). However, the giga-pixeled nature of WSIs poses a great challenge for efficient MIL. Existing studies either do not consider global dependencies among instances, or use approximations such as linear attentions to model the pair-to-pair instance interactions, which inevitably brings performance bottlenecks. To tackle this challenge, we propose a framework named MamMIL for WSI analysis by cooperating the selective structured state space model (i.e., Mamba) with MIL, enabling the modeling of global instance dependencies while maintaining linear complexity. Specifically, considering the irregularity of the tissue regions in WSIs, we represent each WSI as an undirected graph. To address the problem that Mamba can only process 1D sequences, we further propose a topology-aware scanning mechanism to serialize the WSI graphs while preserving the topological relationships among the instances. Finally, in order to further perceive the topological structures among the instances and incorporate short-range feature interactions, we propose an instance aggregation block based on graph neural networks. Experiments show that MamMIL can achieve advanced performance than the state-of-the-art frameworks. The code can be accessed at https://github.com/Vison307/MamMIL.

\end{abstract}


\section{Introduction}
As the gold standard for cancer diagnosis, pathology has been undergoing a new leap from manual observation to digital analysis since the approval of whole slide images (WSIs) \cite{acs2020next}. Accompanied by the growth of deep learning, a new interdisciplinary field, computational pathology, has emerged \cite{song2023artificial}. In computational pathology, deep learning-based models are developed to conduct automatic WSI analysis, which largely frees the work burden for pathologists and alleviates the subjectivity in the diagnosis process \cite{hosseini2024computational}. 

However, WSIs are composed of tens of billions of pixels. Feeding the huge-sized WSIs into deep learning models is often infeasible due to memory limitations in the graphic processing units (GPUs). To this end, researchers have recently paid much attention to multiple instance learning (MIL). In MIL, each WSI is treated as a bag and the small patches split from the WSI are viewed as instances. Mainstream MIL frameworks first extract instance features by feature extractors pre-trained on large-scale datasets. Then, an instance aggregation strategy is designed to obtain a bag feature, which is supervised by the WSI-level label, enabling the model training process. Therefore, feature learning is no longer performed on the giga-pixeled WSIs but conducted on each small instance in MIL, thus alleviating the problem of GPU memory limitation. 

Since the aggregation directly determines the model's performance, many primary MIL frameworks focused on the aggregation strategy \cite{ilse2018attention, shi2020loss, lu2021data, zhang2022dtfd, li2021dual}.
However, all these studies assume the instances are independent of each other. In contrast, tissue interactions are critical in tumor progression. For this reason, several studies constructed MIL frameworks based on graph neural networks (GNN) \cite{li2018graph, chen2021whole}. Nevertheless, GNNs are prone to the over-smoothing issue \cite{chen2020measuring}, which prevents them from being stacked into deep neural networks with multiple layers to model long-range instance interactions. To this end, some studies \cite{shao2021transmil} tried to build MIL methods based on Transformer to model the global instance dependencies. However, the quadratic complexity of self-attention in Transformer impedes the model training process. Hence, Transformer-based MIL methods usually approximate self-attention with linear attention, whose performance is inevitably limited \cite{pmlr-v201-duman-keles23a}. 

To address the above issues, some researchers have tried to apply state space models (SSMs) into MIL due to their ability to model long sequences with linear or near-linear complexity. For example, Fillioux et al. \cite{fillioux2023structured} utilized a diagonal SSM named S4D \cite{gu2022parameterization} for MIL. However, the parameters of S4D are input-invariant, which hinders the model from focusing on the most discriminative instances, leading to inferior performance. Recently, a new kind of SSM, Mamba \cite{gu2023mamba}, has emerged. Compared with previous SSMs, Mamba extends the SSM's parameters to a content-dependent manner. Existing studies verified that Mamba can achieve comparable or better performance than Transformer in various tasks \cite{gu2023mamba,zhu2024vision,liu2024vmamba} with only half of the parameters compared with Transformers. However, several challenges exist in applying Mamba to MIL. Firstly, Mamba is built for one-dimensional (1D) sequences. When flattening the WSIs to 1D sequences as inputs, the loss of topological information inevitably occurs. Secondly, Mamba uses a ``scanning'' strategy to compute the latent states in a unidirectional manner. Although unidirectional scanning is feasible to model sequences such as text and audio with a time series characteristic, it is inefficient for WSIs with pair-wise dependencies. 
\begin{figure*}[!th]
\centering
\includegraphics[width=0.75\textwidth]{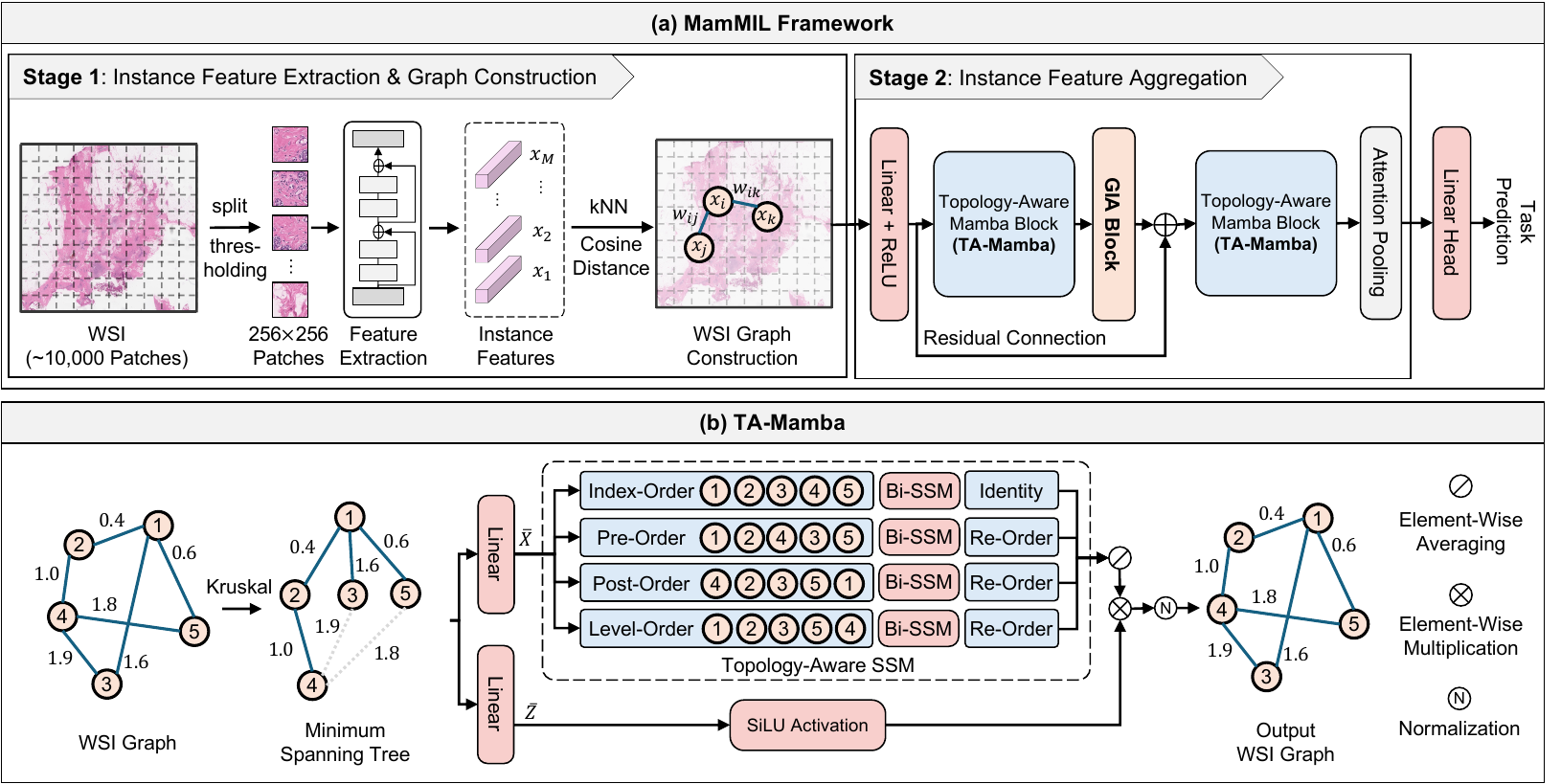}
\caption{An overview of the MamMIL framework, which is composed of an instance feature extraction and graph construction stage, along with an instance feature aggregation stage.\label{fig1}}
\end{figure*}

MambaMIL \cite{yang2024mambamil}, a contemporary study with our work, proposed an SRMamba module to solve the unidirectional scanning problem. However, it did not consider the loss of topological information during the WSI flattening process. In addition, MambaMIL only employed the vanilla Mamba block. Recently, Mamba2 \cite{daotransformers}, an updated version of Mamba, has been proposed with much larger state dimensions, which will further improve the efficiency and performance of MIL. All the above reasons motivate us to propose an MIL framework, dubbed MamMIL. Specifically, the main contributions of this paper are listed as follows.

\begin{itemize}
    \item MamMIL models WSIs as undirected graphs and adopts Mamba2 to achieve global instance interaction. A GNN-based instance aggregation (GIA) block is also designed to capture the instances' local topological relationships.
    \item Considering that Mamba can only model 1D sequences, we develop a topology-aware Mamba (TA-Mamba) block with the minimum spanning tree (MST) to fully preserve the topological information in the 1D instance sequences.
    \item Experiments on two tasks verify that the proposed MamMIL can outperform existing MIL methods, providing a new architecture for future studies.
\end{itemize}

\section{Methodology}
\subsection{Framework Overview}
Fig.~\ref{fig1} shows an overview of MamMIL, which mainly consists of an instance feature extraction and graph construction stage and an instance feature aggregation stage. In the first stage, like in Ref. \cite{lu2021data}, a sliding window with a $256\times256$ size is utilized to split WSIs into small, non-overlapping patches as instances. Next, the cropped patches in the RGB space are transformed into the HSV space, where thresholding is utilized to discard background patches with small tissue areas. Then, all the patches containing large tissue areas are fed into a pre-trained feature extractor. Next, we construct WSI graphs based on the $k$-nearst neighbour (kNN) algorithm, where the cosine distance is adopted to measure the distance between instances.

The second stage utilizes the constructed WSI graphs as the inputs. Firstly, all the instance features are fed into a trainable linear layer with a ReLU activation, and then aggregated by two successive layers of the proposed TA-Mamba block with the GIA block in the middle. Specifically, the GIA block uses a GNN to model the short-distance correlations among adjacent instances, while the TA-Mamba block uses the Mamba2 architecture to perceive global long-distance dependencies. Finally, with an attention-based pooling, a bag-level feature of the WSI is obtained, which is ultimately used to accomplish the WSI analysis task.

\subsection{Instance Feature Extraction and Graph Construction}
Denote a WSI as $X$ and the patches (i.e., instances) split from the WSI as $\{p_i\}_{i=1}^M$, where $p_i \in \mathbb{R}^{256\times 256 \times 3}$ and $M$ is the patch number ($M$ may be different among WSIs). A feature extractor pre-trained on large-scale datasets is employed to extract patch features. After feature extraction, a WSI can be represented by a 1D sequence of instance features, i.e., 
$X = \left[x_1, x_2, \cdots, x_M\right]$.

Considering that the foreground tissue areas of WSIs usually have irregular shapes, structured data representations such as 2D matrices or 1D sequences are unable to fully preserve the topological and spatial relationships among the instances. To this end, this paper uses undirected graphs to represent WSIs. In particular, each instance stands for a node in the graph, and the instance feature is taken as the node feature. The edges between nodes are obtained by the kNN strategy ($k=8$ in this work) in the 2D coordinate space measured by the cosine distance.
Besides, each edge between nodes $i$ and $j$ has a weight $e_{ij}$ , which is measured by the cosine distance of the node features.
In summary, for each WSI, an undirected graph $\mathcal{G} = \{\mathcal{V}, \mathcal{E}, \mathcal{W}\}$ is built, where $\mathcal{V}$, $\mathcal{E}$, and $\mathcal{W}$ represent the set of nodes, edges, and edge weights, respectively.

\subsection{Instance Feature Aggregation}
Considering that the pre-trained feature extractor is frozen, the instance feature aggregation stage begins with a linear layer with a ReLU activation to finetune the instance features. Then, two successive TA-Mamba blocks with a GIA block in the middle are employed to model the instance dependencies. The two blocks are described in detail in the following subsections.

\subsubsection{TA-Mamba Block}
The TA-Mamba block utilizes the SSM in Mamba2 to mine the discriminative dependencies among the vast number of instances with linear complexity. The main purpose of an SSM is to learn a mapping $\text{SSM}: X\rightarrow Y$ from the input instance sequence $X=\{x_i\}_{i=1}^M$ to an output sequence $Y=\{y_i\}_{i=1}^M$ through latent states $\{h_i\}_{i=1}^M$, which are modeled by
\begin{equation}
\label{eq2}
h_i = \bar{\mathbf{A}} h_{i-1} + \bar{\mathbf{B}} x_i, \quad y_i = \bar{\mathbf{C}} h_i.
\end{equation}

In practice, a discretization is required to apply Eq.~\ref{eq2} to deep learning models with discrete inputs and weights. Commonly, $\bar{\mathbf{A}}$, $\bar{\mathbf{B}}$, and $\bar{\mathbf{C}}$ are discretized by the zeroth-order hold rule with a time step $\mathbf{\Delta}$ as
\begin{equation}
\begin{aligned}
\bar{\mathbf{A}} &= \exp(\mathbf{\Delta} \mathbf{A}), \quad \bar{\mathbf{C}}= \mathbf{C}, \\
\bar{\mathbf{B}} &= \left(\mathbf{\Delta} \mathbf{A}\right)^{-1}\left(\exp(\mathbf{\Delta} \mathbf{A})-\mathbf{I}\right) \cdot \mathbf{\Delta}\mathbf{B} \approx \mathbf{\Delta}\mathbf{B}, \\
\end{aligned}
\end{equation}
where $\mathbf{A}$, $\mathbf{B}$, $\mathbf{C}$, and $\mathbf{\Delta}$ are learnable parameters. To enhance the context-perceiving ability, parameters $\mathbf{B}$, $\mathbf{C}$, and $\mathbf{\Delta}$ are correlated with the input sequence $X$ based on three learnable linear projections $s_\mathbf{B}$, $s_\mathbf{C}$, and $s_\mathbf{\Delta}$ by
\begin{equation}
    \begin{aligned}
\mathbf{B} &= s_\mathbf{B}(X), \qquad \mathbf{C} = s_\mathbf{C}(X), \\
\mathbf{\Delta} &= \log(1+\exp(s_{\mathbf{\Delta}}(X) + \mathbf{P}_\mathbf{\Delta})),
\end{aligned}
\end{equation}
where $\mathbf{P}_\mathbf{\Delta}$ represents the learnable parameters for $\mathbf{\Delta}$. 
Compared with vanilla Mamba, Mamba2 simplifies $\mathbf{A}$ from a diagonal structure to a learnable scalar times an identity matrix, and adopts a multi-head mechanism. These modifications enable Mamba2 to fully utilize the acceleration units on modern GPUs with a larger latent state dimension, thereby significantly improving the model's efficiency and performance.

\begin{figure}[!t]
\centering
\includegraphics[width=0.85\linewidth]{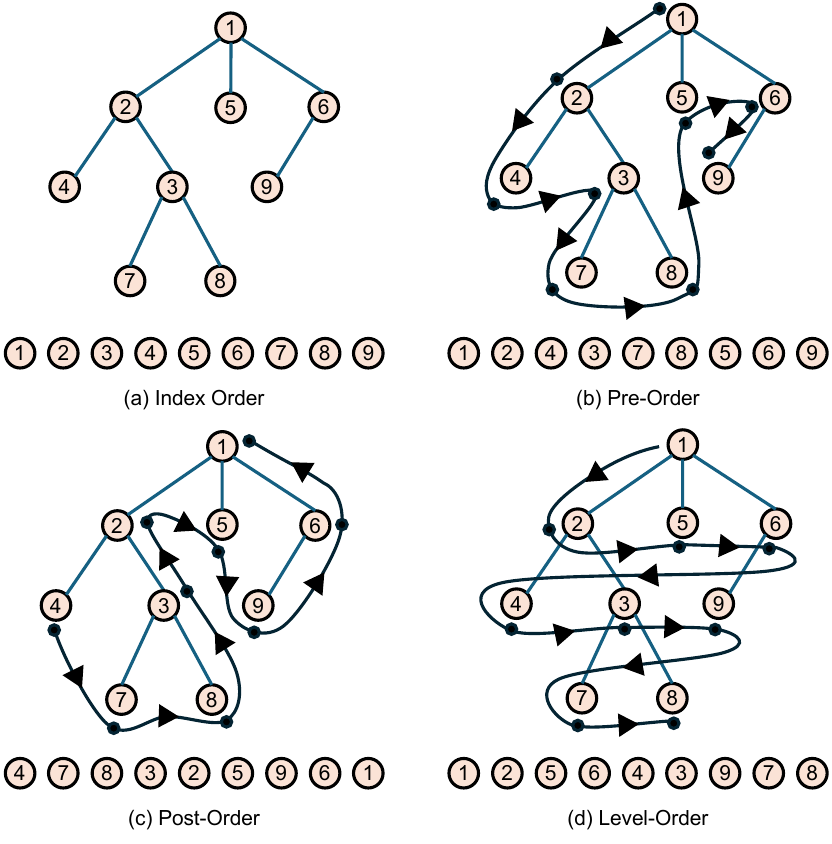}
\caption{Illustrations of the four traversal strategies. The index order traverses the MST using the node index. The pre-order traversal first accesses the root node, and then traverses each sub-tree recursively. The post-order traversal first traverses the sub-trees and accesses the root node at last. The level-order traversal accesses each node from shallower layers to deeper layers. \label{fig2}} 
\end{figure}

However, from Eq.~\ref{eq2}, we can see that the latent state $h_i$ is only related to previous latent states and the current input, making $h_i$ calculated in a unidirectional ``scanning'' manner. Nevertheless, dependencies in any direction may exist in WSIs. 
More importantly, the SSM can only handle 1D sequences, so it cannot directly perform feature aggregation for WSIs represented as graph structures. To this end, we propose a topology-aware scanning mechanism based on MST to serialize WSI graphs to 1D sequences. An MST is a subgraph of $\mathcal{G}$ that ensures the sum of all edge weights is minimal when all nodes are connected. In other words, the MST can be regarded as the most efficient connection scheme of the instances for feature interaction and aggregation. For each WSI graph $\mathcal{G}$, we first use the Kruskal algorithm \cite{kruskal1956shortest} to generate an MST of the graph relative to the edge weights $\mathcal{W}$. Then, we adopt four different traversal strategies to serialize the obtained MST into four 1D sequences to make sure that the topological information in the graph can be preserved as much as possible. Specifically, the original index order of the instances, the pre-order traversal, the post-order traversal, and the level-order traversal of the MST are utilized for serialization, where the pre-order traversal is efficient for describing the structure of the MST, the post-order traversal can quickly access the leaf nodes of the MST, and the level-order traversal describes the hierarchical relationship of the MST. Therefore, taking advantage of these different traversal strategies, the topological structure of the WSI graph can be sufficiently preserved even in the 1D feature sequences, thus facilitating the SSM to capture instance feature interactions and improving the performance of WSI analysis. Illustrations of the four traversal strategies are shown in Fig.~\ref{fig2}. It is worth noting that we randomly select an instance as the root node when traversing the MST.

After obtaining the serialized 1D representations of the WSI graph, we utilize them for feature interaction and fusion. Specifically, for an input sequence $X$, we first feed it into two linear projections to get the outputs $\bar{X}$ and $\bar{Z}$, where $\bar{X}$ is further fed into a proposed topology-aware SSM module for global instance interaction, while $\bar{Z}$ is used to gate the output of the topology-aware SSM module. In the topology-aware SSM module, we first re-order $\bar{X}$ according to the four traversals of the MST to obtain four serialized 1D feature sequences, $\bar{X}_1$, $\bar{X}_2$, $\bar{X}_3$, and $\bar{X}_4$. Then, the four sequences are utilized as inputs for four SSMs. To further avoid the limited receptive field issue caused by SSM's unidirectional scanning, inspired by VIM \cite{zhu2024vision}, we adopt the Bi-SSM mechanism with a bidirectional scanning strategy to model the instance interactions. In Bi-SSM, the original order and the reversed order of a feature sequence are both input into two SSMs, and average pooling is used to aggregate the outputs of both directions. 

Eventually, to fuse the outputs of the four sequences with different traversal orders, we first average the output sequences after re-ordering them to the original index order. Then, a gating mechanism is utilized to obtain the output $X^\prime$ of the TA-Mamba block as
\begin{equation}
    X^\prime = \text{Norm}\bigg(\text{SiLU}(\bar{Z}) \otimes \big(\frac{1}{4} \sum_{i=1}^4 \sigma (\text{Bi-SSM}(\bar{X}_i)) \big)\bigg),
\end{equation}
where $\text{Norm}$ represents a normalization layer, $\otimes$ represents element-wise multiplication, and $\sigma$ represents re-ordering the sequence to index order.

\subsubsection{GNN-based Instance Aggregation Block}
Although TA-Mamba can preserve the topological information of the WSI graphs as much as possible, some structural information is inevitably lost during the serialization process. 
To this end, we further design a GIA block to realize instance interactions among neighbour nodes directly using the WSI graphs.

Specifically, the GIA block adopts the message passing mechanism in DeepGCN \cite{li2019deepgcns} to realize the interaction of neighbour nodes. Technically, for a node $i$ with feature $x_i$, DeepGCN updates the node feature based on the node itself and its first-order neighbour nodes as
\begin{equation}
\label{eq:10}
    x_i^\prime = \text{MLP}(x_i + \mathcal{A}(\{\text{ReLU}(x_j) + \epsilon \vert j \in \mathcal{N}(i) \})),
\end{equation}
where $\text{MLP}$ is a two-layer perceptron, $\epsilon=10^{-7}$ is a small perturbation, and $\mathcal{A}$ is an instance aggregation function. In this work, the softmax function is used for aggregation as
\begin{equation}
    \mathcal{A}(\{x\vert x\in \mathcal{X}\}) = \sum_{\text{all } x} \frac{\exp (x)}{\sum_{\text{all } y \in \mathcal{X}} \exp (y)} \otimes x.
\end{equation}

Finally, with Eq.~\ref{eq:10}, the features of all instances can be aggregated and updated with their neighbour nodes, thus achieving short-distance instance interaction. 

\subsection{Bag Feature Aggregation and WSI Analysis}
After passing the TA-Mamba blocks and the GIA block, all instance features are finally aggregated by an attention-based pooling mechanism \cite{ilse2018attention} to obtain a bag-level representation.
Finally, to achieve the WSI analysis task, the obtained bag representation is fed into a task-specific projection head. For WSI classification, we use cross-entropy loss for model optimization. And for survival analysis, the negative log likelihood loss \cite{chen2021whole} is utilized.

\begin{table}[!t]
\setlength{\tabcolsep}{1pt}
\centering
\caption{Performance comparison over the three datasets. Since the LossAttn framework is built specifically for classification, it cannot be applied to the survival analysis task. We bold the best and underline the second-best results.}
\label{tab:tab1}
\resizebox{\columnwidth}{!}{%
\begin{tabular}{@{}cccccccc@{}}
\toprule
\multirow{2}{*}{Method}                                 & \multicolumn{2}{c}{BRACS-7*}                         & \multicolumn{2}{c}{BRACS-7}                                                & \multicolumn{2}{c}{Camelyon16}                       & TCGA-LUAD                     \\ \cmidrule(l){2-8} 
                                                        & ACC            & \multicolumn{1}{c|}{AUC}            & ACC                       & \multicolumn{1}{c|}{AUC}                       & ACC            & \multicolumn{1}{c|}{AUC}            & C-Index                   \\ \midrule
\multicolumn{8}{c}{\cellcolor[HTML]{EFEFEF}ResNet-50}                                                                                                                                                                                                                          \\ \midrule
\multicolumn{1}{c|}{ABMIL\cite{ilse2018attention}}      & 36.78          & \multicolumn{1}{c|}{72.98}          & ${50.74}_{2.77}$          & \multicolumn{1}{c|}{${79.01}_{3.65}$}          & 80.62          & \multicolumn{1}{c|}{79.92}          & ${60.76}_{4.13}$          \\
\multicolumn{1}{c|}{GatedABMIL\cite{ilse2018attention}} & 39.08          & \multicolumn{1}{c|}{72.92}          & ${50.56}_{2.99}$          & \multicolumn{1}{c|}{${79.45}_{2.85}$}          & 73.64          & \multicolumn{1}{c|}{75.61}          & ${60.60}_{4.24}$          \\
\multicolumn{1}{c|}{CLAM-SB\cite{lu2021data}}           & 37.93          & \multicolumn{1}{c|}{71.86}          & ${51.30}_{3.32}$          & \multicolumn{1}{c|}{${79.09}_{2.88}$}          & 80.62          & \multicolumn{1}{c|}{{\ul 83.67}}    & ${61.98}_{2.21}$          \\
\multicolumn{1}{c|}{CLAM-MB\cite{lu2021data}}           & 40.23          & \multicolumn{1}{c|}{73.81}          & ${54.44}_{5.19}$          & \multicolumn{1}{c|}{${80.51}_{3.73}$}          & 79.84          & \multicolumn{1}{c|}{80.10}          & ${60.63}_{4.89}$          \\
\multicolumn{1}{c|}{DSMIL\cite{li2021dual}}             & 40.23          & \multicolumn{1}{c|}{70.01}          & ${49.63}_{3.95}$          & \multicolumn{1}{c|}{${77.18}_{3.54}$}          & 77.52          & \multicolumn{1}{c|}{75.33}          & ${62.06}_{0.87}$          \\
\multicolumn{1}{c|}{LossAttn\cite{shi2020loss}}         & 28.74          & \multicolumn{1}{c|}{70.57}          & ${51.67}_{2.80}$          & \multicolumn{1}{c|}{${79.00}_{4.36}$}          & 78.29          & \multicolumn{1}{c|}{71.61}          & -                         \\
\multicolumn{1}{c|}{GraphConv\cite{li2018graph}}        & 40.23          & \multicolumn{1}{c|}{70.79}          & ${51.11}_{4.32}$          & \multicolumn{1}{c|}{${77.19}_{5.73}$}          & 80.62          & \multicolumn{1}{c|}{78.44}          & ${62.72}_{1.81}$          \\
\multicolumn{1}{c|}{PatchGCN\cite{chen2021whole}}       & 41.38          & \multicolumn{1}{c|}{71.86}          & {\ul ${55.00}_{3.71}$}    & \multicolumn{1}{c|}{${81.39}_{2.67}$}          & \textbf{82.95} & \multicolumn{1}{c|}{83.14}          & ${62.35}_{2.62}$          \\
\multicolumn{1}{c|}{DTFDMIL\cite{zhang2022dtfd}}        & 39.08          & \multicolumn{1}{c|}{{\ul 76.56}}    & ${53.15}_{4.15}$          & \multicolumn{1}{c|}{${79.75}_{3.18}$}          & 78.29          & \multicolumn{1}{c|}{77.58}          & ${60.96}_{3.21}$          \\
\multicolumn{1}{c|}{TransMIL\cite{shao2021transmil}}    & 36.78          & \multicolumn{1}{c|}{72.25}          & ${45.93}_{4.60}$          & \multicolumn{1}{c|}{${75.83}_{4.15}$}          & 79.84          & \multicolumn{1}{c|}{82.63}          & {\ul ${64.86}_{6.38}$}    \\
\multicolumn{1}{c|}{S4MIL\cite{fillioux2023structured}} & 39.08          & \multicolumn{1}{c|}{73.24}          & ${54.81}_{3.63}$          & \multicolumn{1}{c|}{{\ul ${81.59}_{3.09}$}}    & 80.62          & \multicolumn{1}{c|}{81.79}          & ${64.33}_{2.51}$          \\
\multicolumn{1}{c|}{MambaMIL\cite{yang2024mambamil}}    & {\ul 42.53}    & \multicolumn{1}{c|}{74.14}          & $\mathbf{{56.11}_{4.54}}$ & \multicolumn{1}{c|}{${81.25}_{2.70}$}          & {\ul 82.17}    & \multicolumn{1}{c|}{81.73}          & ${64.37}_{4.06}$          \\ \midrule
\multicolumn{1}{c|}{MamMIL (ours)}                      & \textbf{48.28} & \multicolumn{1}{c|}{\textbf{77.46}} & ${52.96}_{5.75}$          & \multicolumn{1}{c|}{$\mathbf{{81.78}_{2.50}}$} & {\ul 82.17}    & \multicolumn{1}{c|}{\textbf{84.44}} & $\mathbf{{67.18}_{4.02}}$ \\ \midrule
\multicolumn{8}{c}{\cellcolor[HTML]{EFEFEF}VIM}                                                                                                                                                                                                                                \\ \midrule
\multicolumn{1}{c|}{ABMIL\cite{ilse2018attention}}      & 35.63          & \multicolumn{1}{c|}{72.91}          & ${50.19}_{3.84}$          & \multicolumn{1}{c|}{${78.14}_{3.69}$}          & 75.97          & \multicolumn{1}{c|}{74.80}          & ${62.52}_{3.15}$          \\
\multicolumn{1}{c|}{GatedABMIL\cite{ilse2018attention}} & {\ul 37.93}    & \multicolumn{1}{c|}{71.94}          & ${50.19}_{6.65}$          & \multicolumn{1}{c|}{${79.01}_{2.84}$}          & 78.29          & \multicolumn{1}{c|}{{\ul 80.94}}    & ${60.22}_{3.26}$          \\
\multicolumn{1}{c|}{CLAM-SB\cite{lu2021data}}           & 32.18          & \multicolumn{1}{c|}{74.95}          & ${52.96}_{3.81}$          & \multicolumn{1}{c|}{${80.25}_{3.25}$}          & 73.64          & \multicolumn{1}{c|}{71.73}          & ${63.46}_{5.31}$          \\
\multicolumn{1}{c|}{CLAM-MB\cite{lu2021data}}           & 35.63          & \multicolumn{1}{c|}{69.48}          & ${51.11}_{4.70}$          & \multicolumn{1}{c|}{{\ul ${80.26}_{3.09}$}}    & 71.32          & \multicolumn{1}{c|}{75.20}          & ${64.17}_{5.18}$          \\
\multicolumn{1}{c|}{DSMIL\cite{li2021dual}}             & {\ul 37.93}    & \multicolumn{1}{c|}{68.26}          & ${48.89}_{5.81}$          & \multicolumn{1}{c|}{${77.70}_{4.62}$}          & 69.77          & \multicolumn{1}{c|}{58.83}          & ${60.63}_{4.48}$          \\
\multicolumn{1}{c|}{LossAttn\cite{shi2020loss}}         & \textbf{39.08} & \multicolumn{1}{c|}{68.76}          & ${50.00}_{5.17}$          & \multicolumn{1}{c|}{${77.96}_{3.33}$}          & 76.74          & \multicolumn{1}{c|}{72.32}          & -                         \\
\multicolumn{1}{c|}{GraphConv\cite{li2018graph}}        & 33.33          & \multicolumn{1}{c|}{71.43}          & ${49.44}_{3.71}$          & \multicolumn{1}{c|}{${79.11}_{3.93}$}          & {\ul 79.07}    & \multicolumn{1}{c|}{70.59}          & ${63.73}_{5.60}$          \\
\multicolumn{1}{c|}{PatchGCN\cite{chen2021whole}}       & 44.83          & \multicolumn{1}{c|}{{\ul 75.93}}    & {\ul ${53.33}_{5.48}$}    & \multicolumn{1}{c|}{$\mathbf{{80.71}_{3.37}}$} & 78.29          & \multicolumn{1}{c|}{70.20}          & {\ul ${64.23}_{6.55}$}    \\
\multicolumn{1}{c|}{DTFDMIL\cite{zhang2022dtfd}}        & {\ul 37.93}    & \multicolumn{1}{c|}{71.46}          & $\mathbf{{54.26}_{3.42}}$ & \multicolumn{1}{c|}{${79.03}_{3.63}$}          & 70.54          & \multicolumn{1}{c|}{74.46}          & ${63.88}_{5.01}$          \\
\multicolumn{1}{c|}{TransMIL\cite{shao2021transmil}}    & 31.03          & \multicolumn{1}{c|}{66.40}          & ${48.15}_{5.97}$          & \multicolumn{1}{c|}{${77.70}_{3.45}$}          & 73.64          & \multicolumn{1}{c|}{69.03}          & ${60.47}_{4.25}$          \\
\multicolumn{1}{c|}{S4MIL\cite{fillioux2023structured}} & 36.78          & \multicolumn{1}{c|}{69.41}          & ${50.56}_{4.83}$          & \multicolumn{1}{c|}{${77.83}_{2.84}$}          & 71.32          & \multicolumn{1}{c|}{62.88}          & ${62.18}_{1.55}$          \\
\multicolumn{1}{c|}{MambaMIL\cite{yang2024mambamil}}    & 35.63          & \multicolumn{1}{c|}{69.42}          & ${48.15}_{4.46}$          & \multicolumn{1}{c|}{${79.03}_{3.51}$}          & 74.42          & \multicolumn{1}{c|}{69.85}          & ${62.41}_{6.15}$          \\ \midrule
\multicolumn{1}{c|}{MamMIL (ours)}                      & \textbf{39.08} & \multicolumn{1}{c|}{\textbf{77.76}} & ${50.37}_{2.84}$          & \multicolumn{1}{c|}{${79.89}_{2.97}$}          & \textbf{80.62} & \multicolumn{1}{c|}{\textbf{83.67}} & $\mathbf{{65.07}_{2.47}}$ \\ \midrule
\multicolumn{8}{c}{\cellcolor[HTML]{EFEFEF}VMAMBA}                                                                                                                                                                                                                             \\ \midrule
\multicolumn{1}{c|}{ABMIL\cite{ilse2018attention}}      & 42.53          & \multicolumn{1}{c|}{73.91}          & ${52.96}_{4.24}$          & \multicolumn{1}{c|}{${80.26}_{2.42}$}          & 76.74          & \multicolumn{1}{c|}{71.56}          & ${63.63}_{2.53}$          \\
\multicolumn{1}{c|}{GatedABMIL\cite{ilse2018attention}} & 42.53          & \multicolumn{1}{c|}{{\ul 76.97}}    & ${53.89}_{4.79}$          & \multicolumn{1}{c|}{${79.89}_{3.51}$}          & {\ul 80.62}    & \multicolumn{1}{c|}{73.47}          & ${61.88}_{6.66}$          \\
\multicolumn{1}{c|}{CLAM-SB\cite{lu2021data}}           & 36.78          & \multicolumn{1}{c|}{75.86}          & ${54.44}_{4.40}$          & \multicolumn{1}{c|}{${80.43}_{2.75}$}          & 72.87          & \multicolumn{1}{c|}{68.90}          & ${62.32}_{5.96}$          \\
\multicolumn{1}{c|}{CLAM-MB\cite{lu2021data}}           & 40.23          & \multicolumn{1}{c|}{69.08}          & $\mathbf{{56.85}_{3.89}}$ & \multicolumn{1}{c|}{${79.55}_{2.47}$}          & 79.84          & \multicolumn{1}{c|}{{\ul 77.98}}    & ${61.49}_{6.22}$          \\
\multicolumn{1}{c|}{DSMIL\cite{li2021dual}}             & 40.23          & \multicolumn{1}{c|}{74.24}          & ${53.89}_{2.80}$          & \multicolumn{1}{c|}{${79.21}_{2.76}$}          & 67.44          & \multicolumn{1}{c|}{64.74}          & ${60.32}_{6.01}$          \\
\multicolumn{1}{c|}{LossAttn\cite{shi2020loss}}         & 39.08          & \multicolumn{1}{c|}{71.30}          & ${54.44}_{3.90}$          & \multicolumn{1}{c|}{${81.25}_{3.67}$}          & 76.74          & \multicolumn{1}{c|}{73.34}          & -                         \\
\multicolumn{1}{c|}{GraphConv\cite{li2018graph}}        & 41.38          & \multicolumn{1}{c|}{73.01}          & ${54.07}_{4.37}$          & \multicolumn{1}{c|}{${80.45}_{3.22}$}          & 75.97          & \multicolumn{1}{c|}{74.52}          & {\ul ${64.15}_{2.19}$}    \\
\multicolumn{1}{c|}{PatchGCN\cite{chen2021whole}}       & \textbf{45.98} & \multicolumn{1}{c|}{76.25}          & ${52.41}_{2.35}$          & \multicolumn{1}{c|}{${81.58}_{3.26}$}          & {\ul 80.62}    & \multicolumn{1}{c|}{70.13}          & ${62.21}_{5.34}$          \\
\multicolumn{1}{c|}{DTFDMIL\cite{zhang2022dtfd}}        & {\ul 43.68}    & \multicolumn{1}{c|}{75.83}          & ${55.19}_{3.19}$          & \multicolumn{1}{c|}{{\ul ${81.61}_{2.23}$}}    & 77.52          & \multicolumn{1}{c|}{72.12}          & ${61.54}_{4.07}$          \\
\multicolumn{1}{c|}{TransMIL\cite{shao2021transmil}}    & 31.03          & \multicolumn{1}{c|}{73.17}          & ${49.59}_{4.26}$          & \multicolumn{1}{c|}{${77.95}_{2.47}$}          & 68.99          & \multicolumn{1}{c|}{58.98}          & ${61.65}_{5.81}$          \\
\multicolumn{1}{c|}{S4MIL\cite{fillioux2023structured}} & 27.59          & \multicolumn{1}{c|}{66.43}          & ${53.15}_{6.31}$          & \multicolumn{1}{c|}{${78.61}_{4.03}$}          & 73.64          & \multicolumn{1}{c|}{62.78}          & ${61.61}_{3.89}$          \\
\multicolumn{1}{c|}{MambaMIL\cite{yang2024mambamil}}    & 37.93          & \multicolumn{1}{c|}{73.09}          & ${48.70}_{4.69}$          & \multicolumn{1}{c|}{${79.45}_{2.57}$}          & 79.84          & \multicolumn{1}{c|}{76.73}          & ${63.16}_{5.24}$          \\ \midrule
\multicolumn{1}{c|}{MamMIL (ours)}                      & 41.38          & \multicolumn{1}{c|}{\textbf{78.23}} & {\ul ${56.11}_{4.31}$}    & \multicolumn{1}{c|}{$\mathbf{{82.07}_{1.99}}$} & \textbf{82.17} & \multicolumn{1}{c|}{\textbf{81.15}} & $\mathbf{{64.39}_{3.65}}$ \\ \bottomrule
\end{tabular}%
}
\end{table}

\section{Experimental Results}
\subsection{Datasets}
Three public datasets, Camelyon16 \cite{bejnordi2017diagnostic}, BRACS \cite{brancati2022bracs}, and TCGA-LUAD, are utilized for experimental evaluation. Specifically, Camelyon16 and BRACS aim at WSI classification. For Camelyon16, we use the same data split scheme as TransMIL \cite{shao2021transmil}. For BRACS, following Yang et al. \cite{yang2024mambamil}, we use two different settings to evaluate the model's performance. First, the official training, validation, and test WSIs are utilized for performance evaluation, denoted as BRACS-7*. A 10-fold Monte Carlo validation is also adopted to evaluate the proposed MamMIL, denoted as BRACS-7. The TCGA-LUAD dataset is utilized for the survival analysis task. Also following Ref. \cite{yang2024mambamil}, we evaluate the performance with a 5-fold cross-validation.

\subsection{Experimental Settings}
All experiments are done with an RTX 3090 GPU. Codes are implemented with Pytorch 2.1.2. RAdam \cite{liu2021variance} optimizer is utilized with a fixed learning rate of $10^{-4}$ and a weight decay of $0.05$. AUC and C-Index are utilized as the main evaluation metrics. In addition, we report the accuracy with a threshold of 0.5. All results are represented in \%. The number in the lower right corner of a metric represents the standard deviation. Early stopping is employed during training. If the validation loss (for classification) or C-Index (for survival analysis) no longer improves within 20 epochs, the training stops. Each model is trained for up to 250 epochs. The model with the lowest validation loss (for classification) or highest C-Index (for survival analysis) is used for testing. Three backbones, ResNet-50 \cite{he2016deep}, VIM \cite{zhu2024vision}, and VMAMBA \cite{liu2024vmamba}, are utilized for instance feature extraction, which are pre-trained on ImageNet \cite{deng2009imagenet}. Among them, VIM and VMAMBA are also built on Mamba. Therefore, with these two feature extractors, we can build MIL frameworks with a pure Mamba architecture. We extract instance features under a magnification of $20\times$. 

\subsection{Comparison Results}
The comparison results are shown in Table~\ref{tab:tab1}, which proves that the proposed MamMIL exceeds all SOTA methods in both AUC and C-Index over all datasets with ResNet-50 and VMAMBA feature extractors. When adopting VIM as the feature extractor, MamMIL can also achieve the best results over all datasets and metrics except for BRACS-7.
In addition, the results show that compared with our contemporary work, MambaMIL, MamMIL can achieve a performance improvement of more than 1\% over most datasets with different feature extractors. Experimental results also demonstrate that the competing methods' performance varies greatly when applied to different datasets or using different feature extractors. On the contrary, the proposed MamMIL can generally obtain the best AUC and C-Index among all datasets and feature extractors, verifying its robustness. The overall superior performance of MamMIL demonstrates that MIL frameworks built on SSMs and GNNs have great performance potential.

\subsection{Ablation Studies}
To verify the proposed modules, we conduct ablation studies over the BRACS dataset using the ResNet-50 feature extractor. The following two subsections describe the results in detail.
\begin{table}[!t]
\centering
\caption{Ablation study on different scanning strategies. We bold the best and underline the second-best results.}
\label{tab:tab2}
\begin{tabular}{@{}c|cc|cc@{}}
\toprule
\multirow{2}{*}{\begin{tabular}[c]{@{}c@{}}Scanning\\ Strategy\end{tabular}} & \multicolumn{2}{c|}{BRACS-7*}   & \multicolumn{2}{c}{BRACS-7}                           \\ \cmidrule(l){2-5} 
                                                                             & ACC            & AUC            & ACC                       & AUC                       \\ \midrule
Mamba \cite{gu2023mamba}                                                     & {\ul 41.38}    & {\ul 75.71}    & ${52.04}_{4.10}$          & ${81.08}_{2.49}$          \\
BiMamba \cite{zhu2024vision}                                                 & {\ul 41.38}    & 72.47          & ${52.04}_{4.01}$          & {\ul ${81.27}_{2.75}$}    \\
SRMamba \cite{yang2024mambamil}                                              & {\ul 41.38}    & 72.71          & {\ul ${52.22}_{2.72}$}    & ${80.69}_{2.57}$          \\ \midrule
MamMIL (ours)                                                                & \textbf{48.28} & \textbf{77.46} & $\mathbf{{52.96}_{5.75}}$ & $\mathbf{{81.78}_{2.50}}$ \\ \bottomrule
\end{tabular}
\end{table}

\subsubsection{Ablation studies on the Topology-Aware Scanning Mechanism}
To verify the effectiveness of the proposed topology-aware scanning mechanism, we ablate it with three different scanning strategies, including the unidirectional scanning strategy in Mamba, the bidirectional scanning strategy (BiMamba) in VIM. The SRMamba strategy in MambaMIL is also compared. Experimental results in Table \ref{tab:tab2} show that the proposed topology-aware scanning mechanism can achieve the best performance, which verifies that our proposed scanning mechanism can fully preserve the topological information, thus improving the model's performance.

\subsubsection{Ablation studies on the GIA block}

\begin{table}[!t]
\centering
\caption{Ablation study on the GIA block. We bold the best and underline the second-best results.}
\label{tab:tab3}
\begin{tabular}{@{}c|cc|cl@{}}
\toprule
\multirow{2}{*}{\begin{tabular}[c]{@{}c@{}}Aggregation\\ Strategy\end{tabular}}               & \multicolumn{2}{c|}{BRACS-7*}     & \multicolumn{2}{c}{BRACS-7}                                                                   \\ \cmidrule(l){2-5} 
                                & ACC        & AUC             & ACC                                      & \multicolumn{1}{c}{AUC}                       \\ \midrule
w/o GIA                         & {\ul 45.98}    & 75.03          & \multicolumn{1}{l}{$\mathbf{{53.33}_{2.59}}$} & {\ul ${80.49}_{3.55}$}                        \\
w/ PPEG \cite{shao2021transmil} & \textbf{48.28} & {\ul 76.02}    & \multicolumn{1}{l}{${51.11}_{3.33}$}          & ${79.60}_{2.95}$                               \\ \midrule
MamMIL (ours)                   & \textbf{48.28} & \textbf{77.46} & {\ul ${52.96}_{5.75}$}                        & \multicolumn{1}{c}{$\mathbf{{81.78}_{2.50}}$} \\ \bottomrule
\end{tabular}
\end{table}
In this subsection, two ablation methods are employed to verify the effectiveness of the GIA block. First, we remove the GIA block and only use TA-Mamba blocks for long-distance instance interaction. Second, we replace the GIA block with the PPEG module \cite{shao2021transmil}, which uses multi-level convolutions to realize short-range instance interactions. Results shown in Table \ref{tab:tab3} demonstrate that the proposed GIA block can achieve an AUC improvement of over 1\% compared with the second best method. In particular, we can see that the PPEG does not perform better compared with not utilizing the GIA block over BRACS-7. This indicates that the irregular tissue areas in WSIs may make the process in PPEG of reshaping the 1D instance sequences to 2D matrices unreasonable. 
In contrast, our MamMIL uses graphs to represent WSIs, which can fully retain the structural information of the instances to achieve efficient short-range instance feature interaction. Therefore, the proposed GIA block can improve the model's performance.

\section{Conclusion}
Recently, SSMs, especially Mamba, have achieved high performance in modeling long sequences under linear complexity. Considering that WSIs contain tens of thousands of instances, introducing Mamba into WSI analysis can alleviate the performance degradation caused by linear attention approximations in existing Transformer-based MIL frameworks. To this end, we propose MamMIL, which introduces Mamba2 with GNN into the MIL framework. Considering the irregularity of tissue areas, we construct WSIs as undirected graphs. Then, we propose a topology-aware scanning mechanism based on the MST, realizing 1D serialization of WSI graphs while preserving the topological and spatial relationships among the instances. To further perceive the instances' topological structures, we design the GIA block. Experimental results verify that the proposed MamMIL can achieve SOTA performance.

\bibliographystyle{ieeetr}
\bibliography{ref.bib}

\end{document}